# Multi-label natural language processing to identify diagnosis and procedure codes from MIMIC-III inpatient notes [*]

A.K. Bhavani Singh[1], Mounika Guntu[1], Ananth Reddy Bhimireddy[1], Judy W. Gichoya[2], and Saptarshi Purkayastha[1]

[1] Indiana University - Purdue University Indianapolis IN 46202, USA.
{bhavagni, mouguntu, anbhimi, saptpurk}@iu.edu

[2] Emory University, Atlanta GA 30322, USA.
judywawira@emory.edu

**Abstract.** In the United States, 25% or greater than 200 billion dollars of hospital spending accounts for administrative costs that involve services for medical coding and billing. With the increasing number of patient records, manual assignment of the codes performed is overwhelming, time-consuming and error-prone, causing billing errors. Natural language processing can automate the extraction of codes/labels from unstructured clinical notes, which can aid human coders to save time, increase productivity, and verify medical coding errors. Our objective is to identify appropriate diagnosis and procedure codes from clinical notes by performing multi-label classification. We used de-identified data of critical care patients from the MIMIC-III database and subset the data to select the ten (top-10) and fifty (top-50) most common diagnoses and procedures, which covers 47.45% and 74.12% of all admissions respectively. We implemented state-of-the-art Bidirectional Encoder Representations from Transformers (BERT) to fine-tune the language model on 80% of the data and validated on the remaining 20%. The model achieved an overall accuracy of 87.08%, an F1 score of 85.82%, and an AUC of 91.76% for top-10 codes. For the top-50 codes, our model achieved an overall accuracy of 93.76%, an F1 score of 92.24%, and AUC of 91%. When compared to previously published research, our model outperforms in predicting codes from the clinical text. We discuss approaches to generalize the knowledge discovery process of our MIMIC-BERT to other clinical notes. This can help human coders to save time, prevent backlogs, and additional costs due to coding errors.

## 1 Introduction

The electronic health records contain a combination of both structured data such as demographics, laboratory results, medications, etc., and unstructured data such as clinical notes of the patients. Most of the patient data in the EHR

---

[*] This is a shortened version of the Capstone Project that was accepted by the Faculty of Indiana University, in partial fulfillment of the requirements for the degree of Master of Science in Health Informatics.

is present as unstructured text, which is used for labeling patient records. In the United States, diagnosis and procedures are coded from electronic health records using various standard terminologies of which International Classification of diseases (ICD-9-CM) is one of them, and the coded data is used for billing and reimbursement and other secondary uses of data including population health and business analytics. The ICD-9 coding system is widely used in hospitals in the United States to label clinical notes and represents a large corpus of medical terms consisting of over 14,000 diagnoses [9]

Human coders usually peruse through the entire patient record documented during a visit and assign relevant codes from those large code corpora by following a set of rules/guidelines provided by Centers for Medicare and Medicaid Services (CMS) and National Center for Health Statistics (NCHS). This complex process requires significant time and attention from human coders. In situations where only manual coding is used, then the overwhelming volume of generated patient records is difficult to code and leads to errors [10]. Such errors can lead to serious consequences such as overbilling or underbilling by health care organizations resulting in penalties from fraud as well as legal consequences [3]. Incorrect coding can lead to a massive loss in hospital income, which is involved in diagnosis-related groups (DRGs) payment system [12]. It also results in a waste of health care budget. For example, between 2009 and 2012, Medicare had to overpay $35.8 million due to medical coding errors [1]. Automation of code assignment is one of the strategies that has been adopted to tackle the above obstacles for supporting human coders. Computer assisted coding results in increased productivity and improved turnaround time without compromising coding quality [2].

### 1.1 Advances in Named Entity Recognition and Natural Language Processing

Traditional language processing techniques such as word embeddings, NER (Named Entity Recognition) are prominent in providing solutions with text. But these techniques fail to consider the context of the words. Further deep learning models such as RNN's, LSTM's consider the word sequences and provides better results, but when the length of sentence/documents exceeds 20 words the accuracy of the models gradually drops because of memory constraints in the models.

### 1.2 Bidirectional Encoder Representations from Transformers Model (BERT)

This paper proposes a breakthrough NLP transfer learning technique known as Bidirectional Encoder Representations from Transformers (BERT) [4] to automate the task of predicting diagnosis and procedure codes using multi-label text classification from MIMIC-III V1.4 critical care patients' clinical notes [7]. There are two pre-training approaches in NLP, which include a feature-based approach and a fine-tuning based approach. BERT follows a fine-tuning based approach. BERT model was pre-trained on BookCorpus and Wikipedia text for generating two pre-trained models BERT-BASE and BERT-LARGE. The original implementation of the BERT model was in Tensorflow, which is discussed in the subsequent section [4]. We used a PyTorch implementation of BERT for multi-label sequence classification adopting an open-source GitHub repository, which was based on an excellent Huggingface PyTorch repository for BERT [11].

## 2 Methods

### 2.1 Data collection and Preprocessing

Medical Information Mart for Intensive Care (MIMIC) is an openly available database developed by the Massachusetts Institute of Technology (MIT) containing de-identified healthcare information of critical care patients from a world- class teaching hospital [7]. The MIMIC database periodically gets updated with the influx of information, and there are many versions of the database, such as MIMIC II, MIMIC III. We used the updated MIMIC-III V1.4 database released in September 2016, which includes laboratory results, clinical notes, medications, and procedures. MIMIC III V1.4 contains data of 58,000 hospital admissions of over forty thousand patients organized into multiple tables such as "NOTE EVENTS," "CHART EVENTS," and others with columns like admission ID, subject ID, etc. This data was collected between June 2001 and October 2012. These tables are provided as CSV files along with the scripts to load the tables into a database. The selected files contained clinical notes for the patient, their diagnosis and procedures, and the ICD code.

Our task of multi-label classification to predict both diagnosis and procedure codes associated with clinical notes using a single model required us to create a dataset containing clinical notes/note events along with their associated diagnosis and procedure ICD codes. Due to the time and resource constraints and in-view of existing literature which describe the top 10 and 50 diagnosis and procedures [6], we did not use all the note events, diagnosis and procedures contained in the dataset as merging all note events and diagnosis and procedures resulted in hundreds of gigabytes of data. To cover the majority of admissions, we created two different datasets. For the first dataset, we extracted the ten most frequent diagnosis and procedure codes out of a total of 6,985 diagnoses and 2,009 unique codes, respectively. Extracted diagnosis and procedure codes were merged into one table, after which one-hot encoding was performed on both top- 10 diagnosis and top-10 procedure codes, converting them to labels with values as 0 or 1 and their admission IDs as rows. There were two reasons for one-hot encoding for our case: 1) To reduce the output dataset size because when we merged diagnosis and procedures with inner join based on admission ID, there was a duplication of rows as there were multiple rows with same admission ID. 2) Our model expects the input data to be in a one-hot encoded format. As a next step, note events/clinical notes were merged with the diagnosis and procedure table into a single table based on admission. This dataset consists of "27,521" unique admissions (accounting to 1.010829 million rows) out of a total of 58,000 admissions in MIMIC-III.

For the second dataset, we extracted the fifty most frequent diagnosis and procedure codes. One-hot encoding was also performed for top-50 diagnosis and procedure codes before merging with note events. Note events/clinical notes and top-50 diagnosis and procedure codes were merged into a single table based on admission ID. This dataset (top-50) consists of "42,993" unique admissions (accounting to 1.555388 million rows) out of total 58,000 MIMIC-III admissions. Thus, in MIMIC-III V1.4, the top-10 dataset covers 47.45%, and the top-50 dataset covers 74.12% of total admissions. Both datasets were split into train and test sets with an 80:20 ratio using the *sklearn* package. A list of labels for the top-10 and top-50 dataset was passed to the BERT model as a separate text file to define the size of output features in the final layer.

**BERT Model** We used the basic Bert-base-uncased model because it shows the best performance with less computational requirements than the large model. This pre-trained model comes with a tokenizer. It consists of 12 transformer blocks and 12 attention layers with "*gelu*" activation and Adam optimizer with 110M parameters pre-trained on 3300 M words. A dropout probability of 0.1 was applied to all the layers. BERT allows usage of the sequence length of up to 512 tokens. We used the maximum sequence length of BERT, i.e., 512 tokens. We then formatted out data into the expected format as expected by BERT.

## 2.2 Input Representation

**Classification task:** Each example in our training data is assigned a unique ID. BERT tokenizer is used to tokenize the words, and padding is performed to sequences of tokens to keep the sequence length the same in each batch. Input IDs are assigned to each token representing their indices in the defined vocabulary. Input masking is performed by setting 1 to real tokens and 0 to padding tokens.

**Token Embeddings:** BERT base Uncased pre-trained model tokenizes using WordPiece embeddings with 30,522 token vocabulary, which implies BERT al- lowed us to store 30,522 words in the vocabulary. For the words absent in the vocabulary, BERT uses byte pair encoding WordPiece tokenization where a rare word/unknown word is broken down into pieces of sub-words where ultimately, the context of sub-words becomes the context of the rare word. [CLS] tokens are added at the beginning representing the classification task of the token sequences and [SEP] as the sequence separator (especially a pair of sequence texts). The WordPiece tokens are embedded in 768-dimensional vector; in our case, the size then corresponds to (30522, 768) due to the use of 30,522 words vocabulary.

**Positional Embeddings:** BERT uses positional embeddings to indicate that similar words do not need to have the same output representation. The positional embeddings size in our case is represented as (512, 768) as our sequence length is 512. This indicates that, for example, if a sequence of length 512 starts as "Apple is edible," and the other sequence starts as "Apple is famous for iPhone," due to positional embeddings. In this example, BERT differentiates that the meaning of two apples is different.

**Segment Embeddings:** Segment embeddings help BERT differentiate the sequence of tokens in a sentence pair, i.e., which token belongs to which sentence. Thus, this embedding layer contains two vectors representations. Our task did not involve sentence/sequence pairs differentiation.

All these embedding layers were fused into a single shape representing a size of 768, that was passed into BERT's encoder attention layer. BERT model for multiple label sequence classification implements a linear classifier layer on top of the pre-trained BERT model with output features as 20 for the top-10 dataset and as 100 for the top-50 dataset. Binary cross-entropy with logits (BCEWithLogitsLoss) loss function was used to output predictions as independent probabilities.

## 2.3 Fine Tuning

A single classifier layer was used on top of the pre-trained language model of BERT. The complete model was unfrozen to retrain all the parameters instead of just the last layer to produce better results using a training set of the dataset. This took approximately 25 hours for six epochs and 28 hours for four epochs on top-10 and top-50, respectively. We then fine-tuned the model using a learning rate of 3e-5. The fine-tuned model was then used to evaluate both the test sets.

# 3   Results

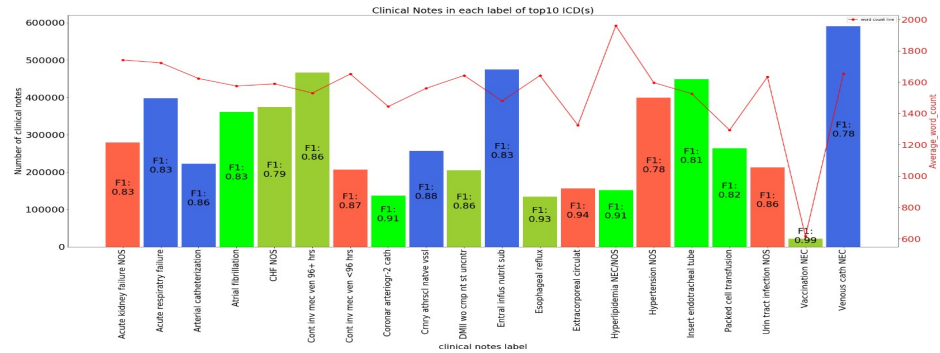

**Fig. 1.** Number of clinical notes for each of top-10 diagnosis and procedure codes

Exploratory data analysis showed that for top-10 data, there is a vast imbalance of clinical notes among the labels ranging from approximately 20,000 notes for "Vaccination NEC" label to around 600,000 clinical notes for "Venous cath NEC" (Figure 1). For top-50, the imbalance of clinical notes ranged from approximately 20,000 for "Coronar arteriogr-1 cath" to around 760,000 for "Venous cath NEC" (Figure 2). Due to the label imbalance in both the datasets, accuracy alone would not be sufficient to define the performance of the model. The average word count for all labels of both top-10 and top-50 datasets ranged from 600 to approximately 2000 (Figure 1, 2 & 3). This word count represents the average number of words required for each label to get their corresponding performance.

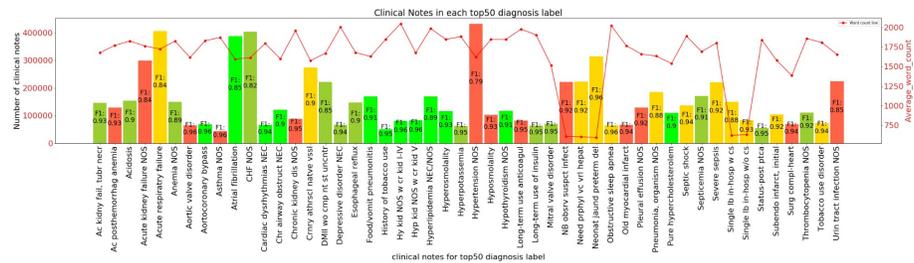

**Fig. 2.** Number of clinical notes for each of top-50 diagnosis codes

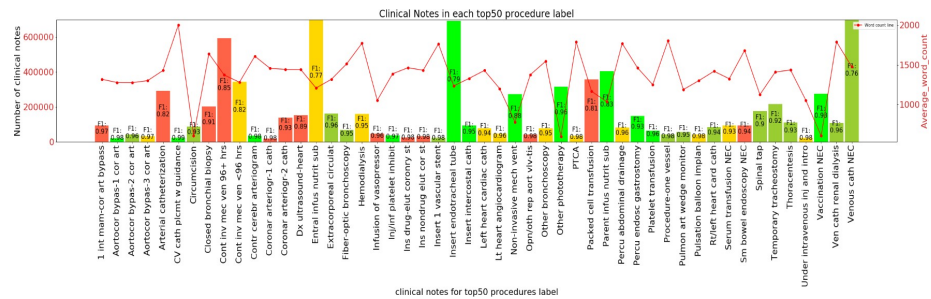

**Fig. 3.** Exploratory Data Analysis: Number of clinical notes for each of top-50 procedure codes

Table 1 represents the evaluation metrics of our model. Figure 4 represents ROC curves for top-10 dataset model. We also had a longer detailed ROC for top-50, but it's too many curves to add to the paper.

We compared our results with published literature using various machine learning and deep learning approaches for assigning codes to clinical notes [5].

**Table 1.** Evaluation of BERT model predictions

| Dataset | Accuracy | Precision (%) | Recall (%) | F1 | AUC |
|---|---|---|---|---|---|
| Top-10 | 87.08 | 87.34 | 86.74 | 85.82 | 91.76 |
| Top-50 | 93.76 | 93.3 | 93.52 | 92.2 | 91 |

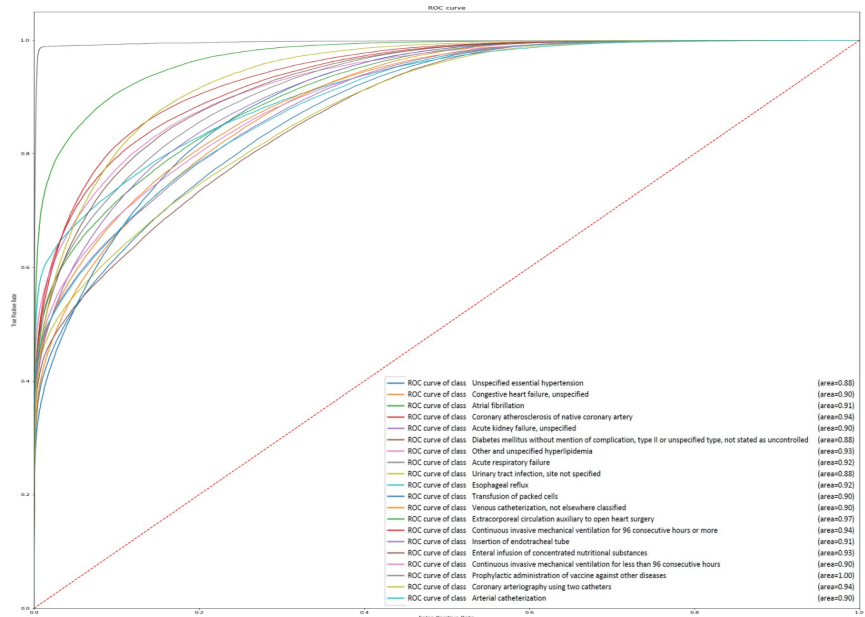

**Fig. 4.** ROC-AUC curves for top-10 diagnosis and procedures codes predictions

[8] [6]. Using various evaluation metrics, we show that our model outputs best results and outperforms other deep learning approaches such as convolutional neural networks and gated recurrent units which is discussed further in the next section.

## 4 Discussion

The evaluation metrics of the top-50 model were higher than top-10, which may be the result of greater text availability for each top-50 label model compared to the top-10. Nonetheless, both models achieve significant results with overall F1 of 85.82 and 92.2, AUC of 91.76 and 91 for top-10 and top-50 predictions, respectively. We evaluated the number of clinical notes for each label along with their F1 score and identified that labels/codes which had few clinical notes had the highest score, such as "vaccination NEC" with 99% F1 score. This indicates that clinical notes which were very specific instead of having more general text are quickly and accurately classified. Models requiring manual feature engineering are labor-intensive, expensive, and not necessarily get the best results. However, since our model was pre-trained and could perform end-to-end learning, we observe better performance with the less computational expense and preprocessing comparative to previous works, as stated in Figure 5. This is a result of using a bidirectional pre-trained language model that resulted in taking advantage of using the contextual information of the English language (in our case) learned from a vast corpus of textual data such as Wikitext. This enabled the model to learn the context of our medical data using the same pre-trained architecture while fine-tuning and classifying. Usage of various evaluation metrics leads to testing the model against imbalanced dataset problem and justified our models' best results by verifying the concordance among those metrics.

**Table 2.** Comparison with published work

| Source | Task | System | F1 | AUC | Precision (%) | Recall (%) | Accuracy |
|---|---|---|---|---|---|---|---|
| Karmakar, 2018 | MIMIC notes classification into Level 1 ICD9 codes | CNN with Attention | 78.2 | - | - | - | - |
| Gehrmann et al., 2018 | Identification of ten patient conditions from text (patient phenotyping) | Convolutional Neural Network (CNN) | 57-97 | 73-100 | 67-100 | 45-95 | - |
| | | cTAKES (RF, full) | 22-72 | 78-99 | 50-81 | 14-80 | - |
| | | cTAKES (LR, full) | 35-83 | 78-98 | 49-93 | 25-79 | - |
| | | cTAKES (RF, filtered) | 37-82 | 85-98 | 53-91 | 29-77 | - |
| | | cTAKES (LR, filtered) | 39-73 | 83-98 | 43-86 | 36-71 | - |
| Huang et al., 2018 | For top-10 codes | Gated recurrent units (GRU) | 69.57 | - | 75.02 | 65.19 | **89.67** |
| | For top-50 codes | GRU | 32.63 | 85.99 | 55.92 | 27.82 | 93.54 |
| This paper | Prediction of ICD and CPT codes from MIMIC-III V1.3 notes for top-10 | BERT model implemented in PyTorch | 77.62-99.49 **(85.82)** | 87.66-99.67 **(91.76)** | 79.76-99.48 **(87.34)** | 78.16-99.49 **(86.74)** | 87.08 |
| | For top-50 multi-label classification | ------------"" ----------- | 76.25-98.54 **(92.2)** | 79.44-99.07 **(91)** | 77.42-98.77 **(93.3)** | 76.51-98.84 **(93.52)** | **93.76** |

We have compared our results with the existing literature corresponding to the ICD code assigned to the clinical notes still because our dataset for a model does not precisely match the literature. For example, a convolutional neural network with attention mechanism implemented to classify 17 Level 1 ICD-9 code [8]. Even though there exists a difference, such as a different number of labels, the reason for comparing is are related to concept extraction from clinical notes. Our current work builds on our previous work of using ULMFiT, which was among the first transfer learning implementation of neural network-based NLP [10]. Our current results outperformed all the models in various evaluation metrics such as F1, AUC, etc., as shown in Table 2, except accuracy of top-10 in comparison to accuracy of Huang et al., top-10 accuracy [6]. The primary reason for our improved results is because we have used a bidirectional pre-trained language model, as stated above. A major limitation of the work is that our model has limited explainability of how it comes up with the codes or its named-entity recognition.

## 5 Conclusion

We demonstrate that with fine tuning based transfer learning using a pretrained bidirectional transformer language model, one can achieve the best results for concept extraction NLP task i.e., diagnosis/procedure codes extraction from clinical notes. BERT classifier shows significant improvement in predicting codes from the clinical text compared to previous research and can help human coders to save time, prevent backlogs, and additional costs due to coding errors. Deployment of the model in pilot trial is important to verify and maximize the effectiveness of the model.

## 6 Future Work

The model can be extended to all ICD/CPT/SNOMED codes instead of restricting to top-10 and top-50. Now we can't interpret the reasons for models' predictions such as the word or the sequence of words that are important to the model. Explainable models can help human coders to quickly identify the reason for the prediction of a code from clinical notes.